\documentclass[10pt,journal,compsoc]{IEEEtran}
\usepackage{lineno,hyperref}
\usepackage{times}
\usepackage{latexsym}
\usepackage{graphicx,subfigure}
\usepackage{epstopdf}
\usepackage{amsmath, amsfonts}
\usepackage{multirow}
\usepackage{upgreek}
\usepackage{bm}
\usepackage{cite}
\usepackage{indentfirst}
\usepackage{graphicx}
\usepackage{color}
\usepackage{amssymb}
\usepackage{algorithm}
\usepackage{algorithmic}
\usepackage{arydshln}
\usepackage{pinyin}

\begin{document}

\title{Chinese Lexical Simplification}

\author{Jipeng~Qiang, Xinyu~Lu, Yun~Li, Yunhao~Yuan, Yang~Shi, and~Xindong~Wu, ~\IEEEmembership{~Fellow,~IEEE,}
\IEEEcompsocitemizethanks{\IEEEcompsocthanksitem J. Qiang, X. Lu, Y. Li, Y. Yuan and Y. Shi are with the Department
of Computer Science, Yangzhou, Jiangsu, China.\protect\\
E-mail: \{jpqiang, 181303216, liyun, yhyuan, shiy\}@yzu.edu.cn
\IEEEcompsocthanksitem X. Wu is with Key Laboratory of Knowledge Engineering with Big Data (Hefei University of Technology), Ministry of Education, Hefei, Anhui, China, and Mininglamp Academy of Sciences, Minininglamp, Beijing, China. \protect\\
E-mail: xwu@hfut.edu.cn
}
}

\markboth{Journal of \LaTeX\ Class Files,~Vol.~14, No.~8, April~2019}%
{Shell \MakeLowercase{\textit{et al.}}: Bare Advanced Demo of IEEEtran.cls for IEEE Computer Society Journals}

\IEEEtitleabstractindextext{%
\begin{abstract}

Lexical simplification has attracted much attention in many languages, which is the process of replacing complex words in a given sentence with simpler alternatives of equivalent meaning. Although the richness of vocabulary in Chinese makes the text very difficult to read for children and non-native speakers, there is no research work for Chinese lexical simplification (CLS) task. To circumvent difficulties in acquiring annotations, we manually create the first benchmark dataset for CLS, which can be used for evaluating the lexical simplification systems automatically. In order to acquire more thorough comparison, we present five different types of methods as baselines to generate substitute candidates for the complex word that include synonym-based approach, word embedding-based approach, pretrained language model-based approach, sememe-based approach, and a hybrid approach. Finally, we design the experimental evaluation of these baselines and discuss their advantages and disadvantages. To our best knowledge, this is the first study for CLS task.

\end{abstract}

}

\maketitle

\IEEEdisplaynontitleabstractindextext

\IEEEpeerreviewmaketitle

\section{Introduction} 

Lexical Simplification (LS) aims at replacing complex words with simpler alternatives without changing the meaning of the sentence, which can help various groups of people, including children \cite{De_belder}, non-native speakers \cite{paetzold2016unsupervised}, people with cognitive disabilities \cite{feng2009automatic,saggion2017automatic}, to understand text better. For example, the sentence "John composed these verses in 1995" could be lexically simplified into "John wrote the poems in 1995". LS task has been applied to different languages, such as English \cite{Carroll1998Practical,glavavs2015simplifying,paetzold2016unsupervised,paetzold2017lexical,gooding-kochmar-2019-recursive,qiang2020lexical}, Japanese\cite{kajiwara2013selecting,kajiwara-yamamoto-2015-evaluation}, Spanish\cite{bott2012can,rello2013frequent}, Swedish\cite{rennes2015tool} and Portuguese\cite{aluisio2010fostering}.

Chinese, the only existing pictographic language in the modern world, is one of the most difficult languages to learn \cite{yang2018makes,wong2017perception}. There are more than 200,000 commonly used words in Chinese that are composed of 5,000 characters.  For example, for a simple Chinese word "\Qi1\zi3" (Wife), there are dozens of equivalent meaning, such as "\Lao3po", "\Po2\niang2", "\Xi2\fu4", "\Nei4\ren2", "\Hai2\ta1\niang2", "\Dui4\xiang4", "\Fu1\ren2", "\Ai4ren", "\Tai4tai" and so on. The complexity and richness words of Chinese text tend to make these people (children, non-native speakers, etc) feel extremely difficult. These suggest that Chinese lexical simplification system is an invaluable tool for improving text accessibility. However, there has been no published work on Chinese lexical simplification so far. Therefore, we focus on the Chinese lexical simplification (CLS) problem in this paper. 

The first challenge of CLS is the lack of human annotation. We first construct a benchmark dataset HanLS for CLS that can be used for both training and evaluation, as well as to accelerate the research on this topic. Firstly, we request two native speakers with teaching experience to give some target words as the list of content words (nouns, verbs, adjectives, and adverbs), and search some sentences containing the target words. Given a sentence and a word to be simplified, we then asked six annotators to give its simpler variants of that word that are appropriate in the context of the sentence.

The second challenge in CLS task is proposing substitutes that are semantically consistent with the original target word and fit in the context but also preserve the sentence's meaning. There have been no published approaches on CLS so far. For providing a comprehensive comparison, we propose five different types of methods as baselines to generate substitutes. (1) Synonym dictionary-based approach: it obtains substitute candidates by picking synonyms from a manually curated lexical dictionary. (2) Word embedding-based approach: it uses the similarity of word embeddings to generate substitute words. (3) Pretrained language model-based approach: we adopt pretrained language model BERT \cite{devlin2018bert} that masks the complex word of the original sentence for feeding into BERT to predict the masked token. (4) Sememe-based approach: we design a word substitution method based on sememes, the minimum semantic units, which can retain more potential valid substitutes for complex words. (5) One hybrid method: we extract candidate substitutions by combining the synonym dictionary and the pretrained language model-based approach. After obtaining the substitute candidates, we utilize the following four features to select the best substitute: language modeling based on BERT, word frequency, word similarity, and Hownet similarity, which respectively capture one aspect of the suitability of the candidate word to replace the complex word.

The contributions of this work are two-fold:

(1) We focus on the Chinese lexical simplification (CLS) task and create manually the first benchmark dataset HanLS for CLS that can be used to evaluate the CLS approaches automatically. 

(2) We propose five different benchmarks for the CLS task, which contains two classic methods (Synonym dictionary and Word embedding) and three latest methods (Pretrained language model, Sememe, and Hybrid). Experimental results show that these baselines (Synonym dictionary, Pretrained language model, and Hybrid) output lexical simplifications that are grammatically correct and semantically appropriate on HanLS.

The dataset and baselines to accelerate the research on this topic are available at https://www.github.com/anonymous.

\section{Related Work}

Lexical simplification (LS) as a sub-task of text simplification focuses to simplify complex words of one sentence with simpler variants. Most current researches are focused on English lexical simplification. We will introduce English LS methods in detail, briefly explain other language LS methods, and finally present some work related to Chinese LS. In addition, we will present the common datasets for each language LS task. All these datasets contain instances that are composed of a sentence, a target complex word, and a set of suitable substitutions provided by humans with respect to their simplicity.

\textbf{English LS and its benchmarks}: The popular lexical simplification approaches were rule-based, in which each rule contains a complex word and its simple synonyms \cite{Lesk:1986:ASD:318723.318728,pavlick2016simple,maddela-xu-2018-word}. Rule-based systems usually identified synonyms from WordNet or other linguistic databases for a predefined set of complex words and selected the "simplest" from these synonyms based on the frequency of word or length of word \cite{devlin1998the,De_belder}. Some LS systems tried to extract rules from parallel corpora \cite{biran2011putting,yatskar2010sake,horn2014learning}. To entirely avoid the requirement of lexical resources or parallel corpora, LS systems based on word embeddings were proposed \cite{glavavs2015simplifying,paetzold2017lexical,gooding-kochmar-2019-recursive}. They extracted the top words as candidate substitutions whose vectors are closer in terms of cosine similarity with the complex word. Pre-training language models \cite{devlin2018bert,DBLP:journals/corr/abs-1904-09223} have attracted wide attention and have shown to be effective for improving many downstream natural language processing tasks. The recent LS methods are based on BERT \cite{qiang2020lexical,zhou-etal-2019-bert} to generate suitable simplifications for complex words. 

There are three widely used datasets for English LS, which are LexMTurk \cite{horn2014learning}, BenchLS \cite{paetzold2016unsupervised} and NNSeval \cite{paetzold2017survey}. LexMTurk is composed of 500 instances annotated by 50 Amazon Mechanical "turkers". BenchLS is composed of 929 instances for English, which is from LexMTurk and LSeval \cite{De_belder}. The LSeval contains 429 instances, in which each complex word was annotated by 46 turkers and 9 Ph.D. students. NNSeval is composed of 239 instances for English, which is a filtered version of BenchLS.  

\textbf{Other language LS}: Most of the other language LS methods are often based on linguistic databases to find simpler candidate substitutes for complex words. The PorSimples project provides an LS method for Brazilian Portuguese, which uses sets of related words provided by the databases Tep 2.0 and PAPEL \cite{aluisio2010fostering}. Bott et al. \cite{bott2012can} use the Spanish OpenTheaurus to find synonyms for complex words in Spanish. Keskisärkkä \cite{keskisarkka2012automatic} used a thesaurus SynLex for the Swedish language to find synonyms for complex words. Kajiwara et al. \cite{kajiwara2013selecting} taken advantage of dictionaries that provide word descriptions. The method extracts candidate substitutions from a complex word's definition. They constructed a dataset from the newswire corpus for the evaluation of Japanese lexical simplification. Afterward, Kodaira et al. \cite{kodaira-etal-2016-controlled} proposed a new controlled and balanced dataset for Japanese lexical simplification with a high correlation with human judgment.

\textbf{Chinese LS}: To our best knowledge, there is no work about Chinese LS. The most relevant work with Chinese LS is Chinese text readability assessment \cite{Liu2017Chinese}. Text readability assessment is used to measure the difficulty level of the given text to assist the selection of suitable reading materials for learners \cite{collinsthompson2014computational}. Automatic text readability measures are composed of formula-based method and classification method using various features, including word features, sentence features, etc. When the difficulty level of the text is obtained, the next step is to simplify the original text for reducing the difficulty of the text and meeting the needs of different users. However, Chinese LS task receives little attention, and we cannot obtain publicly available methods and datasets. Therefore, in this paper, we will first construct a Chinese LS dataset for evaluation, and propose some different LS systems to simplify Chinese sentence.

\section{A Dataset}

After referring the construction of existing English and Japanese lexical simplification datasets, we create a dataset HanLS for Chinese lexical simplification task annotated by three undergraduates and three graduate students. These students are all native Chinese speakers. We follow these steps below. 

(1) \textbf{ Extracting sentences}: We define complex words as "High Level" words in the worldwide popular Chinese HSK vocabulary \cite{Zhao2003Some}. The 600 high-level words (nouns, verbs, adjectives, and adverbs) are chosen by two native speakers with teaching experience based on their experience and intuition. Our aim is to create a balanced corpus and control sentences to have only one complex word. Then, sentences that include a complex word are randomly extracted from these two sources: Modern Chinese corpus of the State Language Commission and Chinese translation corpus \footnote{https://github.com/brightmart/nlp\_chinese\_corpus}. Following previous work, 10 sentences including each complex word are collected. Annotators chose one sentence for each complex word under each POS tag by controlling the number of complex words in each sentence. 

(2) \textbf{Providing substitutes}: Simplification candidates were collected from five native speakers. For each instance, the annotators wrote substitutes that did not change the sense of the sentence. When providing a substitute, an annotator could refer to a dictionary but was not supposed to ask the other annotators for an opinion. When an annotator could not think of a paraphrase, they were permitted to supply no entry. These annotators ranked their provided several substitutes for the complex word according to how simple they were in contexts. 

(3) \textbf{Merging All Annotations}: All annotations were merged into one dataset by averaging the annotations from all annotators. An example from this dataset is explained below. Given one example, we suppose it has one substitute $x$. When the following rankings (1,2,2,4,1) were obtained from five annotators, the average rank of $x$ was 2. The final integrated ranking for each instance is obtained by rearranging the average ranks of these substitutes in the ascending order. 

The merging dataset was evaluated by a new annotator. The annotator rated a substitute as inappropriate based on the following two criteria: i) A substitute is inappropriate if the sentence becomes unnatural after replacing the target word; ii) A substitute is inappropriate if the meaning of the sentence is changed after replacing the target word. Finally, the dataset has 524 instances where each instance has an average of 8.51 substitutes, denoted as HanLS. The complex words in HanLS contain nouns 166, verbs 160, adjectives 134, and adverbs 64, which are composed of one character 9, two characters 472, three characters 13, and four characters 30, respectively. Figure 1 shows an example of the dataset. Here, the complex word has 9 substitutes and we only show four of them.

\begin{figure}
  \centering
  \includegraphics[width=75mm]{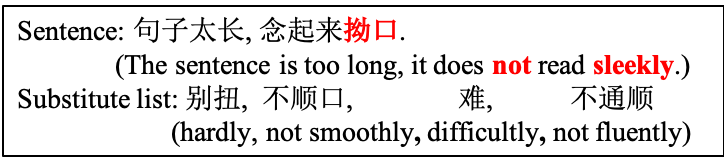}
  \caption{An example of annotation in the dataset HanLS. The word with red color is the complex word.}
  \label{Fig1}
\end{figure}

\section{Baselines}

Following the steps of English lexical simplification \cite{glavavs2015simplifying,paetzold2017survey}, Chinese lexical simplification system also includes the following three steps: complex word identification, substitution generation, and substitution ranking. In the complex word identification (CWI) step, the goal is to select the words in a given sentence which should be simplified. We perform CWI implicitly during other steps of the pipeline. We consider all words in a sentence to be targets for simplification, but during the simplification process we discard substitutions that, when applied $(w_i \rightarrow w_i)$, replace a word $w_i$ with a more complex alternative. The aim of Substitution Generation (SG) is to produce substitute candidates for the complex words. We present five different methods for SG. Giving substitute candidates of the complex word, the Substitution Ranking (SR) of the lexical simplification is to decide which one of the candidate substitutions that fits the context of the complex word is the simplest. We adopt four high-quality features to rank the substitutes. The structure of our framework is shown in Figure 2.
 
\begin{figure}
  \centering
  \includegraphics[width=85mm]{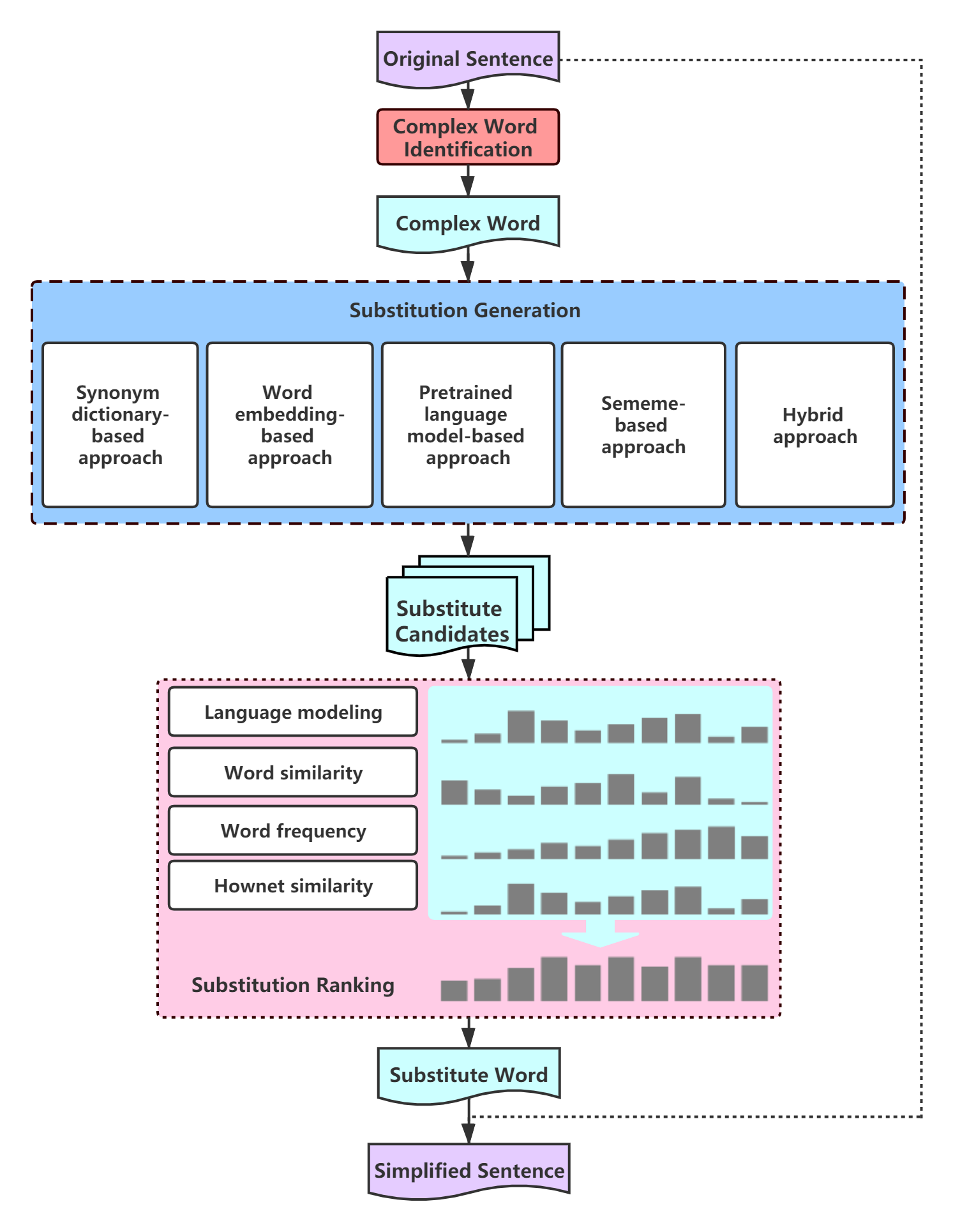}
  \caption{Chinese lexical simplification framework.}
  \label{Fig2}
\end{figure}

\subsection{Substitution Generation}
 
 An ideal SG strategy will be able to find all words that can replace a given target complex word in all contexts in which it may appear. For providing a comprehensive comparison, we provide five different types of approaches to generate substitutes for Chinese LS task and discuss their advantages and disadvantages. 

\textbf{ (1) Synonym dictionary-based approach}: Most LS approaches \cite{bott2012can,aluisio2010fostering} utilized synonym dictionary for SG, e.g., WordNet for English and OpenThesaurus for Spanish. For Chinese SG, we choose a synonym thesaurus HIT-Cilin \cite{mei1996tongyici} for generating substitutes, which contains 77,371 distinct words. The advantage of the method is simple and easy to implement. Besides constructing a synonym dictionary that is expensive and time-consuming, it is impossible to cover all the words.

\textbf{(2) Word embedding-based approach}: Word embedding-based approaches \cite{paetzold2016unsupervised} was used for English SG, which first obtains the vector representation for each word from the pretrained word embedding model and extracts the top $k$ words as substitutes whose embeddings vector has the highest cosine similarity with the vector of the complex word. Here, we use the pretrained Chinese word vectors \footnote{https://github.com/Embedding/Chinese-Word-Vectors} using Word2Vector algorithm \cite{mikolov2013efficient}, and extract the top 10 words as substitutes. The advantage of the method is the pretrained word embedding model is easily accessible because it only needs an ordinary large amount of text corpus. The substitute candidates contain not only similar words, but also highly related words and words with opposite meanings.

\textbf{(3) Pretrained language-model based approach}: Recent English LS method \cite{qiang2020lexical,zhou-etal-2019-bert} adopted pretrained language model BERT to produce substitutes. BERT is a bi-directional language model trained by two training objectives: masked language modeling (MLM) and next sentence prediction (NSP). Unlike a traditional language modeling objective of predicting the next word in a sequence given the history, MLM predicts missing tokens in a sequence given its left and right context. Different from English LS task, we cannot directly utilize Chinese pretrained BERT model for Chinese SG. Because English has a natural space as a separator, we only mask the word $w$ of the sentence $S$ using one special symbol "[MASK]" to obtain the probability distribution of the vocabulary corresponding to the mask word. 

In Chinese, a word is composed of one or more characters. For one complex word is composed of four characters, the possible substitutes may be one character, two characters, three characters, and four characters. We need to use different numbers of [MASK] symbol to replace the complex word. Therefore, predicting the [MASK] symbols is not only a cloze task but also a generating task.

Specifically, for one complex word, we use less than or equal to the number of [MASK] symbols to replace it and combine all the results as substitutes. The original sentence $S$ replaced with [MASK] symbols is denoted as $S'$. Considering BERT is adept at dealing with sentence pairs, we feed the sentence pair $\{S, S'\}$ into the BERT. Suppose that $S'$ contains two [MASK] symbols. We first obtain the top $n$ candidate characters for the first [MASK] symbol. For each candidate character, we replace the corresponding [MASK] of $S'$ into the candidate character and feed the new sentence pair $\{S, S'\}$ into BERT for obtaining the top $n$ candidate characters for the second [MASK] symbol. We filter these words that are not in the Modern Chinese Word List \cite{yuming2007green} after combining the second candidate characters with the first candidate character. This scheme achieves better results than simultaneously predicting the candidate characters for the two [MASK] symbols. 

This method is the only one that making use of the wider context when generating substitute candidates. In all experiments, we use BERT-Base, Chinese pretrained model \footnote{https://huggingface.co/bert-base-chinese}.

\textbf{ (4) Sememe-based approach}: The meaning of a word can be represented by the composition of its sememes, where sememe is defined as the minimum indivisible semantic unit of human languages defined by linguists \cite{bloomfield1926set}. Sememes have been successfully used for many NLP tasks including semantic composition \cite{qietal2019modeling}, pretrained language model \cite{zhang-etal-2020-enhancing}, etc. This is the first attempt to apply sememe for lexical simplification. 

In practical NLP applications, Sememe knowledge bases are built based on sememes, in which Hownet is the most famous one\cite{dong2006hownet}. In Hownet, the sememes of a word can accurately describe the meaning of the word. Therefore, the words owning the same sememe annotations should share the same meanings, and they can act as the substitute candidates for each other. In our sememe-based method, a word $w$ can be substituted by another word $w*$ only if one of $w$'s senses has the same sememe annotations as one of $w*$'s senses.

Compared with the word embedding and language model-based substitution methods, sememe-based approach cannot generate many inappropriate substitutes, such as antonyms and semantically related but not similar words. Compared with the synonym-based method, sememe-based method generates more substitute words. 

\textbf{(5) Hybrid approach}: We design a simple hybrid approach for Chinese SG, which combines the synonym dictionary-based approach and the pretrained language model-based approach. Specifically, if the complex word is included in HIT-Cilin synonym dictionary, we use the synonym dictionary-based approach to generate substitutes, else we use the pretrained language model-based approach.

For the above substitution generation methods, in our experiments, we filter these substitutes that are not in the dictionary (Modern Chinese Word List).

\subsection{Substitution Ranking}

 We choose four different features for SR. Each of the features captures one aspect of the suitability of the candidate word to replace the complex word. In addition to the word frequency, word similarity, and language model features commonly used in other language LS methods, we consider one additional high-quality Hownet similarity feature. We compute various rankings according to their scores for each of the features.

\textbf{(1) Language modeling}: The aim of the feature is to evaluate the fluency of substitute in a given sentence. We do not choose traditional n-gram language modeling, and we choose the pretrained language model BERT to compute the probability of a sentence or sequence of words. Because of the MLM of BERT, we cannot directly compute the probability of a sentence using BERT. Let $W=w_{-m},...,w_{-1},w,w_1,...,w_m$ be the context of the original word $w$. We adopt a new strategy to compute the likelihood of $W$. We first replace the original word $w$ with the substitution candidate. We then mask one word of $W$ from front to back and feed into Bert to compute the cross-entropy loss of the mask word. Finally, we rank all substitute candidates based on the average loss of $W$. The lower the loss, the substitute candidate is a good substitution for the original word. We use as context a symmetric window of size five around the complex word.

\textbf{(2) Word similarity}: We obtain the vector representation of each word using the pretrained word embedding model, and compute the similarity between the complex word and each substitute. The higher the similarity value, the higher the ranking.

\textbf{(3) Word Frequency}: Frequency-based substitute ranking strategy is one of the most popular choices by English lexical simplification. In general, the more frequency a word is used, the most familiar it is to readers. In this work, we adopt the word frequency which is calculated from one big corpus \footnote{https://github.com/liangqi/chinese-frequency-word-list} which contains more than 2.5 hundred million characters. We test many word frequency files from different corpora, and this one we adopted is proved to be the best one. 

\textbf{(4) Hownet similarity}: In addition to the word similarity using word embeddings, we choose a new word similarity method based on Hownet, which has been proved that it has a good performance in antonym and synonym similarity calculation for Chinese words\cite{liu2002word}. Hownet-based similarity based on the sememes computes the similarity between the complex word and the substitutes, which provides a good complimentary for the following situation. When the substitute candidates are antonyms and semantically related but not similar words, the two features (language model and word similarity) probably lose their effectiveness.

\subsection{CLS System}

The overall CLS system is shown in Algorithm 1. We try to simplify each content word (noun, verb, adjectives, and adverb) in the sentence $S$ (line 1). We first choose one substitution generation method from the above five methods to generate the substitutes for the complex word $w$ (line 2). Afterward, we compute various rankings for each of the simplification candidates using each of the features, and then scores each candidate by averaging all its rankings (lines 4-12). We choose the top two substitutes with the average rank scores over all features (line 13). If the first substitute is not the complex word $w$, we will replace the complex word $w$ into the first substitute (lines 14-15). Otherwise, if the first substitute is the complex word $w$, we will choose the second substitute only if the second substitute has a higher frequency than the complex word (lines 17-18). 

\begin{algorithm}[tb]
\caption{Simplify(sentence $S$)}
\label{alg:algorithm}
\begin{algorithmic}[1] 
\FOR{ each content word $w\in S$ }
\STATE $subs$ $\leftarrow$ Substitution$\_$Generation($s$,$w$)
\STATE $all\_ranks$ $\leftarrow$ $\varnothing$
\FOR{ each ranking feature $f$ }
\STATE $scores$ $\leftarrow \varnothing$
\FOR{ each $sc\in scs$ }
\STATE $scores$ $\leftarrow$ $scores$ $\cup$ $f(sc)$
\ENDFOR
\STATE $rank$ $\leftarrow$ $rank\_numbers(scores)$
\STATE $all\_ranks$ $\leftarrow$ $all\_ranks$ $\cup$ $rank$
\ENDFOR
\STATE $avg\_rank$ $\leftarrow$ $average(all\_ranks)$
\STATE $first, second$  $\leftarrow$ TopTwo$_{sc}(avg\_rank)$
\IF{$first$ $\neq$ $w$}
\STATE Replace($S$,$w$,$first$)
\ELSE
\IF{$word\_freq(second$)$>$$word\_freq(w)$}
\STATE Replace($S$,$w$,$second$)
\ENDIF
\ENDIF
\ENDFOR
\end{algorithmic}
\end{algorithm}

\section{Experiments}

We design experiments to answer the following three questions:

\textbf{Q1. The quality of the created Chinese lexical dataset HanLS:} Is the results of manual evaluation consistent with that of annotated dataset HanLS?

\textbf{Q2. The difference of the proposed five substitution generation methods: } The evaluation metrics from previous English LS task are used to verify the effectiveness of these different SG methods on HanLS.

\textbf{Q3. The factors of affecting the CLS system: } We conduct experiments on HanLS to verify the influence of some key parameters (substitution generation methods and substitution ranking features) on the whole CLS system.

Here, the proposed CLS methods are called as synonym dictionary-based method (\textbf{Synonym}), word embedding-based approach (\textbf{Embedding}), pretrained language model-based approach (\textbf{Pretrained}), sememe-based approach (\textbf{Sememe}) and a hybrid approach (\textbf{Hybrid}).

\subsection{Evaluation of the quality of the dataset HanLS}

\begin{table}
\centering
\begin{tabular}{|l|l|}
\hline
 & Embed \ Sememe  \ Pretr  \ \ Hybrid \ \ Synon\\
\hline

Changed& \ \ \ 472 \ \ \ \ \  \  442  \ \ \ \ \  \ \ \ 503 \ \ \ \ \ \ \  470  \ \ \ \ \ \ \ 379 \\
Manual& 0.708 $<$ 0.799 $<$ 0.827 $<$ 0.864 $<$ 0.917  \\
Auto& 0.623 $<$ 0.692 $<$ 0.716 $<$ 0.785 $<$ 0.854 \\
\hline
\end{tabular}
\caption{ The comparative results of manual evaluation and automatic evaluation for HanLS. Embed, Pretr and Synon are short for Embedding, Pretrained and Synonym. }
\label{HumanResults}
\end{table}

Considering the richness of Chinese vocabulary, we plan to verify the comprehensive of the annotated reasonable substitutes in HanLS. We design an experiment to compare the difference between the results of the manual evaluation and the results of automatic evaluation using the annotated substitutes. We adopt the following metrics. It should be noted that we only consider these instances in which the complex word is changed by the system, rather than all instances in HanLS, because we cannot evaluate the annotated substitutes for these instances with no replacement.

\textbf{Changed}: The number of instances in which the complex word is changed by the system.

\textbf{Manual}: The proportion of instances in which the complex word is replaced correctly by manual evaluation.

\textbf{Auto}: The proportion of instances in which the complex word is replaced with any of the substitutes in the dataset.

The results are shown in Table \ref{HumanResults}. From the ranking order of these five methods, we can see that the results of the manual evaluation are in accordance with the results of automatic evaluation. The average proportion of instances in which the results of the manual evaluation is the same as the results of the automatic evaluation is above 85\%. Synonym achieves the best values using Manual and Auto. But it only generates the substitutes for 379 instances, which also means that many complex words are replaced by the original word itself. We conclude that HanLS is a high-quality dataset in which the annotated substitutes are reasonable and comprehensive. Below, we will give a detailed comparison of the baselines we proposed using HanLS.

\subsection{Evaluation of substitution generation}

\begin{table}
\centering
\begin{tabular}{|l|cccc|}
\hline
SG methods &  Potential & Precision & Recall & F1 \\
\hline
Synonym& 81.49 & 40.68  & \textbf{27.42} & \textbf{32.76}  \\
Embedding& 72.14 & 19.70  & 35.36 & 25.30 \\
Pretrained& 88.93 & 31.41  & 26.23 & 28.59 \\
Sememe& 72.14 & 30.76 & 13.24 & 18.51  \\
Hybrid& \textbf{90.46} & \textbf{42.90} & 26.40 &  32.69 \\
\hline
\end{tabular}
\caption{Substitution generation evaluation results (\%). }
\label{SGResults}
\end{table}

We use the following four metrics from the previous English LS task \cite{paetzold2017survey,qiang2020lexical} to evaluate the performance of the SG method.

\textbf{Potential}: The proportion of instances for which at least one of the substitutes generated is in the gold-standard.

\textbf{Precision}: The proportion of generated substitute candidates that are in the annotated substitutes.

\textbf{Recall}: The proportion of annotated substitutes that are included in the generated substitution candidates.

\textbf{F1}: The harmonic mean between Precision and Recall.

The results are shown in Table 2. We can see that the two methods (Synonym and Pretrained) are more effective than the two methods (Embeddings and Sememe). Embedding has the lowest Precision value, because the generated substitutes contain many semantically related but not similar words. For Sememe-based method, it generates dozens or even hundreds of substitutes for many instances, which results in the poorest Recall value. Synonym-based method is a simple but powerful method, which can be easily understood and deployed to different languages. But both Synonym and Sememe have a big limitation that is their coverage. For example, we can find that many common used words do not occur in this dictionary, e.g., "\yuan2\zhu4(Assistance)", "\xing2\nang2(Luggage)" and "\ke1\po4(break up)" for Synonym dictionary, "\xian3\you3(rare)", "\chun2\shu3(purely)" and "\huang1\man2(wild)" for Sememe. Pretrained method without relying on linguistic databases offers impressive results, mainly because it considers the context of the complex word when generating substitute candidates. The hybrid method offers the highest Potential and Precision. 

Overall, Pretrained and Hybrid methods offer the best Potential. Pretrained provides a good balance Precision and Recall using only pretrained language model trained over raw text. Considering the nature of the strategies discussed and the results of our benchmark, it is likely to conclude that the combination of different strategies can create competitive substitution generators.

\subsection{System Evaluation and Ablation Study}

\begin{table*}
\centering
\begin{tabular}{|l|cc|cc|cc|cc|cc|cc|}
\hline
& \multicolumn{2}{|c|}{Synonym}  & \multicolumn{2}{|c|}{Embedding}  & \multicolumn{2}{|c|}{Pretrained} & \multicolumn{2}{|c|}{Sememe} & \multicolumn{2}{|c|}{Hybrid}\\ \hline
 &  PRE & ACC &  PRE & ACC & PRE & ACC & PRE & ACC & PRE & ACC  \\ \hline
w/o Language& 72.33 & 62.60  & 58.78  & 54.39 & 69.27 & 65.27 & 58.40 & 57.44 & 77.86 & 67.56   \\ 
w/o Similarity& 70.99 & 64.12 & \textbf{60.88} & \textbf{56.49} & 66.03 & 63.93 & 49.24 & 48.28 & 76.72 & 69.27 \\
w/o Frequency & 70.42 & 48.66 & 56.68 & 52.29 & 72.90 & 62.60  & 55.92 & 54.96 & 76.15 & 53.82 \\
w/o Hownet & 73.47 & 63.93 & 57.44 & 53.05 & 67.75 & 64.69 & 59.35 & 58.40 & 79.77 & 69.66 \\
\hline
Full & \textbf{74.43} & \textbf{64.69} & 60.50 & 56.11  & \textbf{73.09} & \textbf{68.70} & \textbf{59.35} & \textbf{58.40}& \textbf{80.73} & \textbf{70.42} \\
\hline
\end{tabular}
\caption{ Full pipeline results and Ablation study results of the ranking features.}
\label{rankingfeatures}
\end{table*}

Besides, we use these two previous metrics to evaluate the performance of the full pipeline. To determine the importance of each ranking feature, we make an ablation study by removing one feature in turn. The results are presented in Table \ref{rankingfeatures}. 

\textbf{Precision (PRE)}: The proportion with which the replacement of the original word is either the original word itself or is in the gold standard.

\textbf{Accuracy (ACC)}: The proportion with which the replacement of the original word is not the original word and is in the gold standard.

We first analyze the influence of each feature for the performance of each lexical simplification method. We can see that all approaches combining all four features achieve the best results, except Similarity feature for Embedding, which means all features have a positive effect. Embedding removing Similarity feature produces almost identical results with Embedding combining all features. Word Embedding-based approach has already used word embeddings to generate substitute candidates which lead to Similarity feature that has no effect on substitution ranking.  
 
Then, we compare the full pipeline results of the five methods. Hybrid attains the highest Accuracy and Precision. Pretrained also achieves the satisfactory experiment results. Although the results of Synonym are very encouraging, the main drawback of Synonym is its coverage. The best English LS method \cite{qiang2020lexical} on its benchmark dataset NNSeval obtained a Precision score of 0.526 and an Accuracy score of 0.436. Compare with English LS task, we can find that the three approaches (Synonym, Pretrained, and Hybrid) on Chinese LS task can be served as strong baselines.

\subsection{Error Analysis}

\begin{table*}
\centering
\begin{tabular}{|l|cccc|c|}
\hline
 &  2 & 3 & 4 & 5 & 1 \\
\hline
Synonym& 97(18\%)& 51(10\%) & 185(35\%)  & 50(10\%)  & 289(55\%)  \\
Embedding&  146(28\%)& 111(21\%) & 230(44\%)  & 85(16\%)  & 209(40\%)  \\
Pretrained&  58(11\%) & \textbf{30(6\%)} & 164 (31\%)  & \textbf{23(4\%)}   & \textbf{337(64\%)}  \\
Sememe&  146(28\%)& 56(11\%) & 218(42\%)  & 39(7\%)  & 267(51\%)    \\
Hybrid&  \textbf{50(10\%)} & 57(11\%) & \textbf{155(30\%)}  & 51(10\%)  & 318(61\%)   \\
\hline
\end{tabular}
\caption{Error categorisation results of the baselines. }
\label{Error}
\end{table*}

In this subsection, we analyze all proposed approaches to understand the sources of its errors. We use PLUMBErr tool \cite{paetzold2017lexical} to assess all steps taken by LS systems, and identify five types of errors.

\begin{description}
\item[1)] No error during simplification.
\item[2)] No candidate substitutions are produced.
\item[3)] No simpler candidates are produced.
\item[4)] Replacement compromises the sentence’s grammaticality or meaning.
\item[5)] Replacement does not simplify the word.
\end{description}

Errors of type 2 and 3 are made during Substitution Generation, and error 4 and 5 during Substitution Ranking. Table \ref{Error} shows the count and proportion (in brackets) of instances in HanLS in which each error was made. It shows that Pretrained correctly simplifies the largest number of problems while making the fewest errors of type 3 and 5. However, it can be noticed that Pretrained makes many errors of 4. Hybrid makes the fewest error of type 2 and 4. Embedding making the most mistakes for each step is the worst method compared with other methods. By analyzing the output produced after each step, we found that this is caused by producing many semantically related but not similar words as substitute candidates. Synonym and Sememe make few errors of type 3 and 5, but they make many errors of type 2 and 4. They are based on linguistic databases, in which many complex words cannot be found in the databases. Overall, the results are in accordance with the conclusions of the above experiments.

\section{Conclusion}

In this paper, we manually built a dataset for the performance evaluation of Chinese lexical simplification (CLS) system automatically. We proposed five different methods to generate the substitute candidates and introduced four high-quality features to rank the substitute candidates. Experiment results have shown that synonym-based approach, pretrained language model-based approach, and hybrid method achieved better results. We believe the proposed CLS systems will serve as strong baselines and the created dataset can accelerate the research on this topic for future research. Despite some initial positive results on a difficult task, we note that the performance of CLS system can be affected by substitution generation and substitution ranking. In the future, we will incorporate some prior knowledge into pretrained language model for CLS.

\section*{Acknowledgement}

This research is partially supported by the National Natural Science Foundation of China under grants 61703362 and 91746209; the National Key Research and Development Program of China under grant 2016YFB1000900; the Program for Changjiang Scholars and Innovative Research Team in University (PCSIRT) of the Ministry of Education, China, under grant IRT17R32; and the Natural Science Foundation of Jiangsu Province of China under grant BK20170513.

\bibliographystyle{IEEEtran}
\bibliography{CLS.ref}

\end{document}